\documentclass[pdflatex,sn-mathphys-num]{sn-jnl}

\usepackage{graphicx,subfigure}
\usepackage{multirow}%
\usepackage{amsmath,amssymb,amsfonts}%
\usepackage{amsthm}%
\usepackage{mathrsfs}%
\usepackage[title]{appendix}%
\usepackage{xcolor}%
\usepackage{textcomp}%
\usepackage{manyfoot}%
\usepackage{booktabs}%
\usepackage{algorithm}%
\usepackage{algorithmicx}%
\usepackage{algpseudocode}%
\usepackage{listings}%
\usepackage{caption}
\newcommand{\E}{\mathbb{E}}
\newcommand{\R}{\mathbb{R}}
\DeclareMathOperator*{\argmin}{arg\,min}

\theoremstyle{thmstyleone}%
\newtheorem{theorem}{Theorem}
\theoremstyle{thmstyletwo}%
\newtheorem{remark}{Remark}%

\theoremstyle{thmstylethree}%

\begin{document}


\title[Article Title]{Explainable Learning Rate Regimes for Stochastic Optimization}


\author*[1]{\fnm{Zhuang} \sur{Yang}}\email{zhuangyng@163.com}

%

\affil*[1]{\orgdiv{School of Computer Science and Technology}, \orgname{Soochow University}, \orgaddress{\street{Ganjiang}, \city{Suzhou}, \postcode{215006}, \state{Jiangsu}, \country{China}}}

%


\abstract{Modern machine learning is trained by stochastic gradient descent (SGD), whose performance critically depends on how the learning rate (LR) is adjusted and decreased over time. Yet existing LR regimes may be intricate, or need to tune one or more additional hyper-parameters manually whose bottlenecks include huge computational expenditure, time and power in practice. This work, in a natural and direct manner, clarifies how LR should be updated automatically only according to the intrinsic variation of stochastic gradients. An explainable LR regime by leveraging stochastic second-order algorithms is developed, behaving a similar pattern to heuristic algorithms but implemented simply without any parameter tuning requirement, where it is of an automatic procedure that LR should increase (decrease) as the norm of stochastic gradients decreases (increases). The resulting LR regime shows its efficiency, robustness, and scalability in different classical stochastic algorithms, containing SGD, SGDM, and SIGNSGD, on machine learning tasks.

}




\maketitle


Stochastic gradient descent (SGD) \cite{sclocchi2024different,tang2023neural,lecun2015deep,bottou2012stochastic} and its variants, including but not limited to SGD with momentum (SGDM) \cite{sutskever2013importance}, SIGNSGD \cite{jang2024rethinking}, AdaGrad \cite{duchi2011adaptive}, RMSProp \cite{tieleman2017divide}, Adam \cite{kingma2014adam}, AdamW \cite{zhou2024towards}, etc., have been the first choice to minimize the loss and train machine learning and also deep learning models since the appearance of back-propagation. More specifically, during recent decades, these SGD-type algorithms have formed an important foundation of several open source frameworks, such as PyTorch, TensorFlow, PaddlePaddle, etc., and achieved remarkable success in solving applications of machine learning and deep learning \cite{van2024hardware,Xie2025LoCo,benjamin2022efficient,jordan2015machine}.  When minimizing an empirical loss on a training dataset with size $m$, SGD accelerates by estimating the loss gradient employing a mini-batch of datasets randomly per step \cite{cotter2011better}. Such operation in SGD significantly reduces the computational cost in contrast to the gradient descent (GD) method, where the latter may be prohibited due to large-scale models.

SGD is only of an extremely simple algorithmic framework, but it does not affect its remarkable breakthrough in areas of computer vision, natural language processing, automatic control, electrical engineering, management science, etc. Nevertheless, obtaining a good performance with SGD and its variants usually needs some manual adjustment of the initial value of the learning rate (LR) for each model and each problem, and designing a careful LR regime. More generally, tremendous studies follow the guideline that a lower computational cost at each step requires SGD using a decreasing LR sequence to guarantee the convergence of stochastic algorithms \cite{wang2022automatic,wang2021convergence}. On the one hand, a decreasing LR sequence further slows the convergence of the algorithm. On the other hand, although existing some certain guidelines, there is no detailed criterion to show us how we should decease the learning rate over time just through several obscure hints. With a fast decreasing LR sequence, it moves too cautiously; with a tardily decreasing LR sequence, it generates erratic and unreliable progress.

Currently, the rules of updating the LR sequence may be too intractable to be implemented, or less effective in practical applications. That how to select an appropriate LR sequence for SGD and its variants has attracted significant attention and increased research interest. We briefly review several crucial learning rate schemes below.

%

\begin{enumerate}

\item One typical selection for LR, $\alpha_t$, in stochastic optimization algorithms is to decrease LR as fast as $\Theta(t^{-c})$ for some constant $c$. For instance, the LR sequence broadly used in SGD is decreased according to a scheme of the form $\alpha_t=\alpha_0(1+\gamma t)^{-1}$, where $\alpha_0>0$ is an initial LR, $\gamma>0$ is a well-calibrated constant, and $t$ denotes the number of iterations. Under using the form of $\alpha(t)=O(t^{-1})$, originally developed by \cite{robbins1951stochastic}, the study \cite{bach2011non} explored how hyper-parameters like $\alpha_0$ and $\gamma$ impact the convergence speed. Such the choice of the LR sequence is a key factor to guarantee asymptotically optimal convergence of stochastic algorithms. More specifically, the lower boundary of stochastic optimization algorithms was over all possible sequence of decreasing LRs \cite{nguyen2019tight,braun2017lower}.

\item In most studies, a well-calibrated constant LR is used \cite{dieuleveut2020bridging,lakshminarayanan2018linear,
ghadimi2016mini} for stochastic optimization algorithms because it is easy to implement in practice and may be tuned to perform well for specific instances. More specifically, a large number of modern stochastic optimization algorithms, such as SVRG (with $\alpha=\frac{0.1}{H}$, here, $H$ denotes the Lipschitz constant) \cite{johnson2013accelerating}, SARAH with ($\alpha=O(\frac{1}{H})$) \cite{nguyen2017sarah}, SPIDER (with $\alpha=\frac{\varepsilon}{Hn_0}$) \cite{fang2018spider}, etc., establish a convergence guarantee in the case of using a constant LR. However, it has been confirmed by several studies that problems with a fixed LR will demonstrate a lag if there is a rapid change in the mean of the observed data \cite{george2006adaptive}. In addition, a constant LR, used in stochastic optimization algorithms, may depend on the property of the model itself, such as the Lipschitz constant, or the convexity parameter, etc., where those hyper-parameters are hard to be derived in practice.
\item Several heuristic algorithms have been developed for obtaining an adaptive LR sequence in accordance with the current data (see \cite{plakhov2009stochastic,george2006adaptive} for review). As an instance, for a multiplicative LR modification rule \cite{plakhov2009stochastic}: relying on the current data, the LR is multiplied either by a constant less than 1, i.e., $\alpha_t=d \alpha_{t-1}$ ($d<1$) if $\nabla f_i(x_t)\nabla f_i(x_{t-1})\leq 0$, or by a constant greater than 1, i.e., $\alpha_t=u \alpha_{t-1}$ ($u >1$) if $\nabla f_i(x_t)\nabla f_i(x_{t-1})> 0$. As pointed out by the work \cite{plakhov2009stochastic}, the choices of $d=0.95$ and $u=1.01$ are for the best possible precision $\varepsilon=10^{-3}$ with approximately 200 iterations, while the choices of $d=0.95$ and $u=1.03$ are for the best possible precision $\varepsilon=10^{-4}$ with 300 iterations. In addition, the work \cite{saridis2007learning} decremented the LR sequence for errors of opposite signs and incremented the LR sequence if the consecutive errors are of the same sign such that the estimate moved closer to the optimal parameter value.

\item There is an increasing interest in SGD and its variants with automatically tuning learning rates (LRs) by the exponential smoothing technique, such as AdaGrad, RMSProp, Adam, AdamW, AMSGRAD \cite{reddi2018convergence}, mentioned above, where they scale coordinates of stochastic gradients by squaring roots of some types of averaging of the squared coordinates in historical gradients. The generic iterative scheme of these adaptive gradient methods can be formulated as
\begin{equation}
\label{f-1}
\begin{split}
&m_t=\phi_t(g_1, \cdots, g_t) \\
&V_t=\psi_t(g_1, \cdots, g_t)\\
&x_{t+1}=x_t-\alpha_t m_t/\sqrt{V_t},
\end{split}
\end{equation}
where $g_i=\nabla f_i(x_i)$ denotes the gradient of the loss function $f_i(x)$ at $x_i$. As several special cases, when setting $\phi_t(g_1, \cdots, g_t)=g_t$, $\psi_t(g_1, \cdots, g_t)=\frac{\mathrm{diag}(\sum_{i=1}^tg_i^2)}{t}$, and $\alpha_t=\alpha/\sqrt{t}$, the iterative scheme \eqref{f-1} becomes AdaGrad. In contrast, when adopting $\phi_t(g_1, \cdots, g_t)=(1-\beta_1)\sum_{i=1}^t\beta_1^{t-i}g_i$ and $\psi_t(g_1, \cdots, g_t)=(1-\beta_2)\mathrm{diag}(\sum_{i=1}^t\beta_2^{t-i}g_i^2)$,
where $\beta_1$, $\beta_2 \in [0, 1)$, the iterative scheme \eqref{f-1}  turns out to be classical Adam. A value of $\beta_1=0.9$ and $\beta_2=0.999$ is typically suggested in practice. In recent years, many studies have made massive efforts on the convergence of these adaptive gradient algorithms.


\item Other popular LR regimes, used for SGD and its variants, evaluate the LR adaptively according to a function of the errors in the predictions or estimates, including the stochastic Polyak learning rate, the line search technique, the Barzilai-Borwein (BB) learning rate, the hyper-gradient gradient descent, and meta-learning procedure (see \cite{yang2023adaptive} for review). In deterministic optimization background, when minimizing an upper bound, $Q(\alpha)=\|x_t-x_*\|^2-2\alpha[f(x_t)-f(x_*)]+\alpha^2\|g_t\|^2$, on the distance of the iteration $x_{t+1}$ to the optimal solution $x_*$: $\|x_{t+1}-x_*\|^2\leq Q(\alpha)$, we obtain the Polyak learning rate under deterministic backgrounds,
\begin{align}
\label{f-2}
\alpha_t=\argmin_{\alpha} Q(\alpha)=\frac{f(x_t)-f(x_*)}{\|g_t\|^2},
\end{align}
 where $f(x_*)$ denotes the optimal function value. The work \cite{loizou2021stochastic} extended the Polyak learning rate into stochastic settings with the form
 \begin{align}
 \label{f-3}
 \alpha_t=\frac{f_i(x_t)-f_i(x_*)}{c\|\nabla f_i(x_t)\|^2},
 \end{align}
 where $c>0$ can be set theoretically based on the property of the function under study. In addition, one can obtain the BB learning rate \cite{barzilai1988two}, namely,
 \begin{align}
 \label{f-4}
  \alpha_t=\frac{\|s_t\|^2}{\langle s_t, y_t\rangle},
   \end{align}
   by minimizing the residual of the secant equation $\|B_ts_t-y_t\|^2$ with $B_t=(1/\eta_t)I$, $s_t=x_t-x_{t-1}$, and $y_k=\nabla f(x_t)-\nabla f(x_{t-1})$, where $I$ denotes an identity matrix. In stochastic backgrounds, the BB-type learning rate \cite{Yang2018Random} is working with the form
   \begin{align}
   \label{f-5}
   \alpha_t=\frac{\gamma}{b_H}\cdot \frac{\|x_t-x_{t-1}\|^2}{\langle \nabla f_{S_H}(x_t)-\nabla f_{S_H}(x_{t-1}), x_t-x_{t-1} \rangle},
   \end{align}
    where $\nabla f_{S_H}(x_t)=\frac{1}{b_H}\sum_{i \in S_H}\nabla f_i(x_t)$, $S_H \subset \{1, 2, \cdots, m\}$, and $\gamma>0$ controls the convergence speed of the algorithm.

\end{enumerate}

Many popular LR regimes for stochastic optimization algorithms have been discussed above. It is not difficult to see that the existing LR regimes may be hard to implement in practice, or require to tune one or more additional hyper-parameters manually whose bottlenecks include huge computational expenditure, time and power for large-scale learning problems. Without loss of the generality, it is worth pointing out that most of LR regimes may depend on some certain information, such as gradient information, the value of the loss function, the iterative information, etc., see \eqref{f-1}-\eqref{f-5}. Naturally, we are very curious about whether there exists a certain relationship between the LR and this information. Further, whether the existing updating rules of LRs are interpretable and universal for different applications. Unfortunately, it seems that we cannot obtain such information from \eqref{f-1}-\eqref{f-5}. Therefore, the interpretability and universality of the existing LR regimes for practical applications are relatively poor.



In all existing LR regimes, heuristic algorithms may offer a better interpretability of when and how the LR sequence decreases or increases for the algorithms over time, but requiring tuning of additional hyper-parameters, which may be intractable for large-scale problems where several thousands of parameters need to be estimated. In the work \cite{schaul2013no}, the authors relied on local gradient variations across samples to determine LRs, where such the method automatically mimicked the procedure of an annealing schedule. Nevertheless, the introduction of curvature information may prevent the work \cite{schaul2013no} for large-scale models. Also, the universality of the work \cite{schaul2013no} in other stochastic optimization algorithms, such as SGDM, SIGNSGD, etc., is still underexplored.

To address this challenge, this work clearly ascertains how the variation of gradient information influences the LR sequence in a simple and direct way, thus leading to an explainable learning rate regime for stochastic optimization algorithms. Further, this work will provide answers for the following two questions to better understand the motivation and contributions of this work.

\begin{enumerate}
\item[\bf{Q1:}] \emph{How does the variation of stochastic gradients influence LRs at any time for different practical applications?}

\item[\bf{Q2:}] \emph{What form of the LR regime can accurately reflect such certain relationships between stochastic gradients and LRs?}

\end{enumerate}

Summarily, for the first question, we will elucidate that the norm of stochastic gradients decreases to allow an increasing LR sequence for stochastic optimization algorithms, while the norm of stochastic gradients increases to demand a decreasing LR sequence for stochastic optimization algorithms. Such the idea of updating the LR sequence may be regarded as a kind of heuristic algorithm. Although such a fact looks very simple, there is no study to distinctly quantify this simple relationship. Therefore, for the second question, motivated by the stochastic quasi-Newton (SQN) algorithm \cite{lin2022explicit,mokhtari2020stochastic,byrd2016stochastic}, we will provide a simple mathematical formula to accurately reflect the relationship we claim.



\section{Results}\label{sec2}

We here empirically answer the above two questions. For the first question, we separately plot the details of the variation of stochastic gradients and LRs (where the LR sequence is computed by the novel LR regime) in figures over time, respectively. For the second question, we validate the efficacy of adaptive LR regimes in improving different stochastic optimization algorithms. More specifically, three common stochastic optimization algorithms, containing SGD, SGDM, and SIGNSGD, are equipped with the resulting LR scheme. For clarity, the resulting novel stochastic optimization algorithms are termed as  A-SGD, A-SGDM, and A-SIGNSGD, respectively. Further, we compare the results between the resulting algorithms and their original algorithms that use both a diminishing LR sequence and different constant LRs for training neural networks (with one fully connected hidden layer of 300 nodes and ten softmax output nodes; sigmoid activation) on the $MNIST$ dataset.

\begin{figure}[!hptb]
\centering
\includegraphics[width=15.5cm]{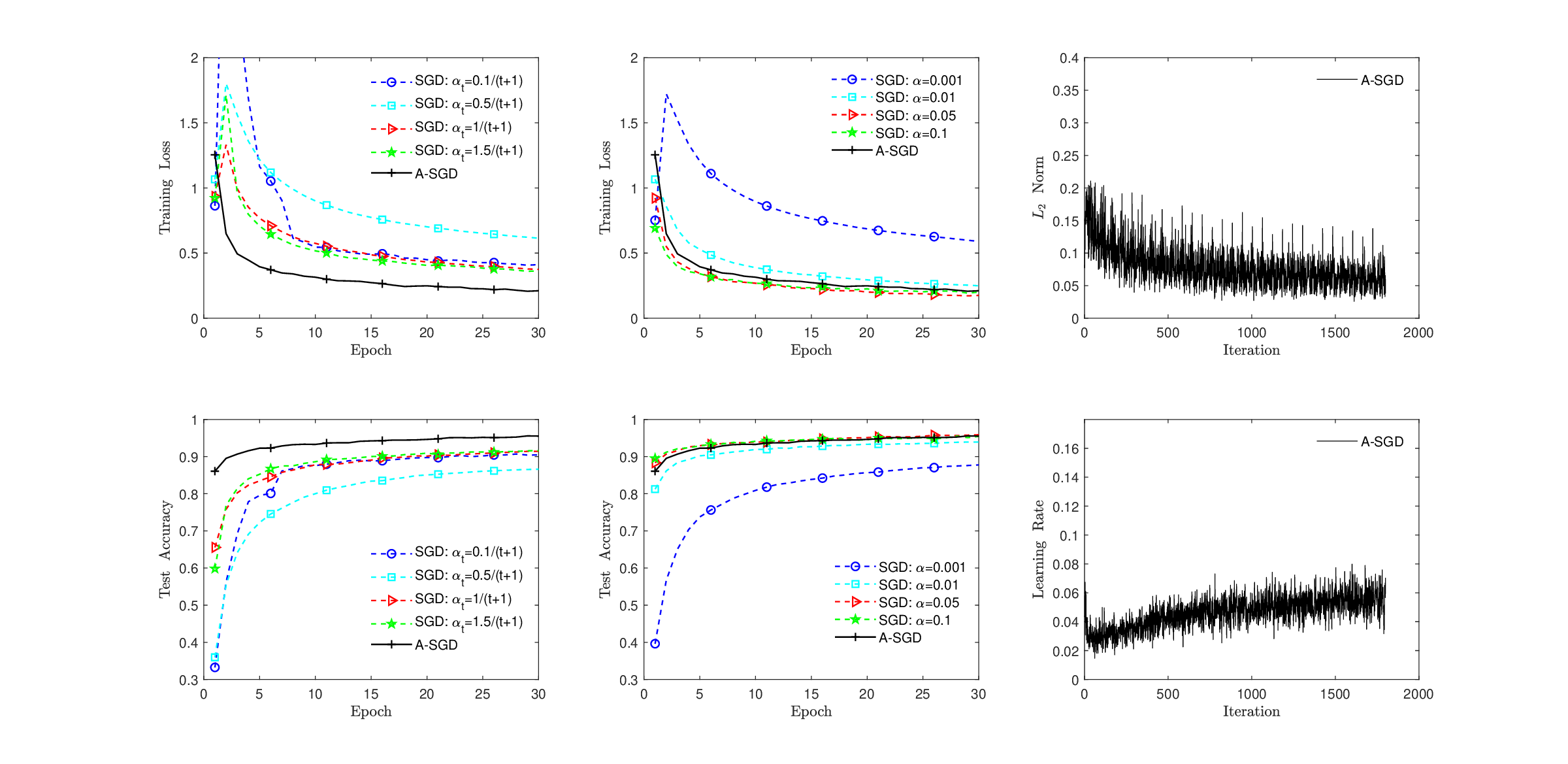}
\caption{MLP results on MNIST for A-SGD. The figure on the first and second columns plots the training loss comparison and classification accuracy of A-SGD and SGD with a diminishing LR sequence and different constant LRs in 30 epochs, respectively. The figure on the third column plots $\|\nabla F_{S_H}(x_t)\|$ and LRs of A-SGD every 10 iterations. The numbers in the legends are the LR, used in SGD. }\label{asgd-mlp}
\end{figure}

\begin{figure}[!hptb]
\centering
\includegraphics[width=15.5cm]{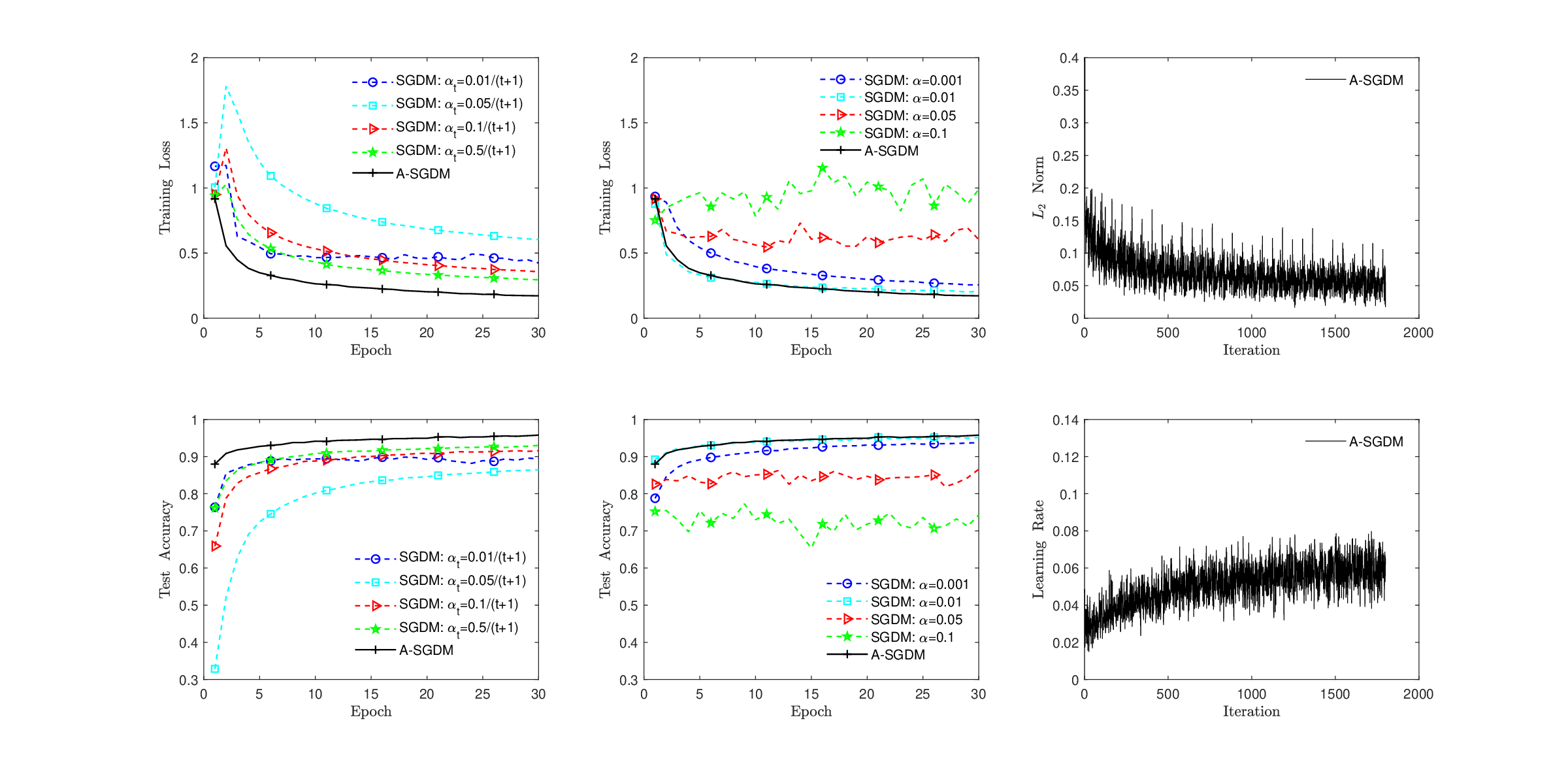}
\caption{MLP results on MNIST for A-SGDM. The figure on the first and second columns plots the training loss comparison and classification accuracy of A-SGDM and SGDM with a diminishing LR sequence and different constant LRs in 30 epochs, respectively. The figure on the third column plots $\|\nabla F_{S_H}(x_t)\|$ and LRs of A-SGDM every 10 iterations. The numbers in the legends are the LR, used in SGDM. }\label{asgdm-mlp}
\end{figure}

\begin{figure}[!hptb]
\centering
\includegraphics[width=15.5cm]{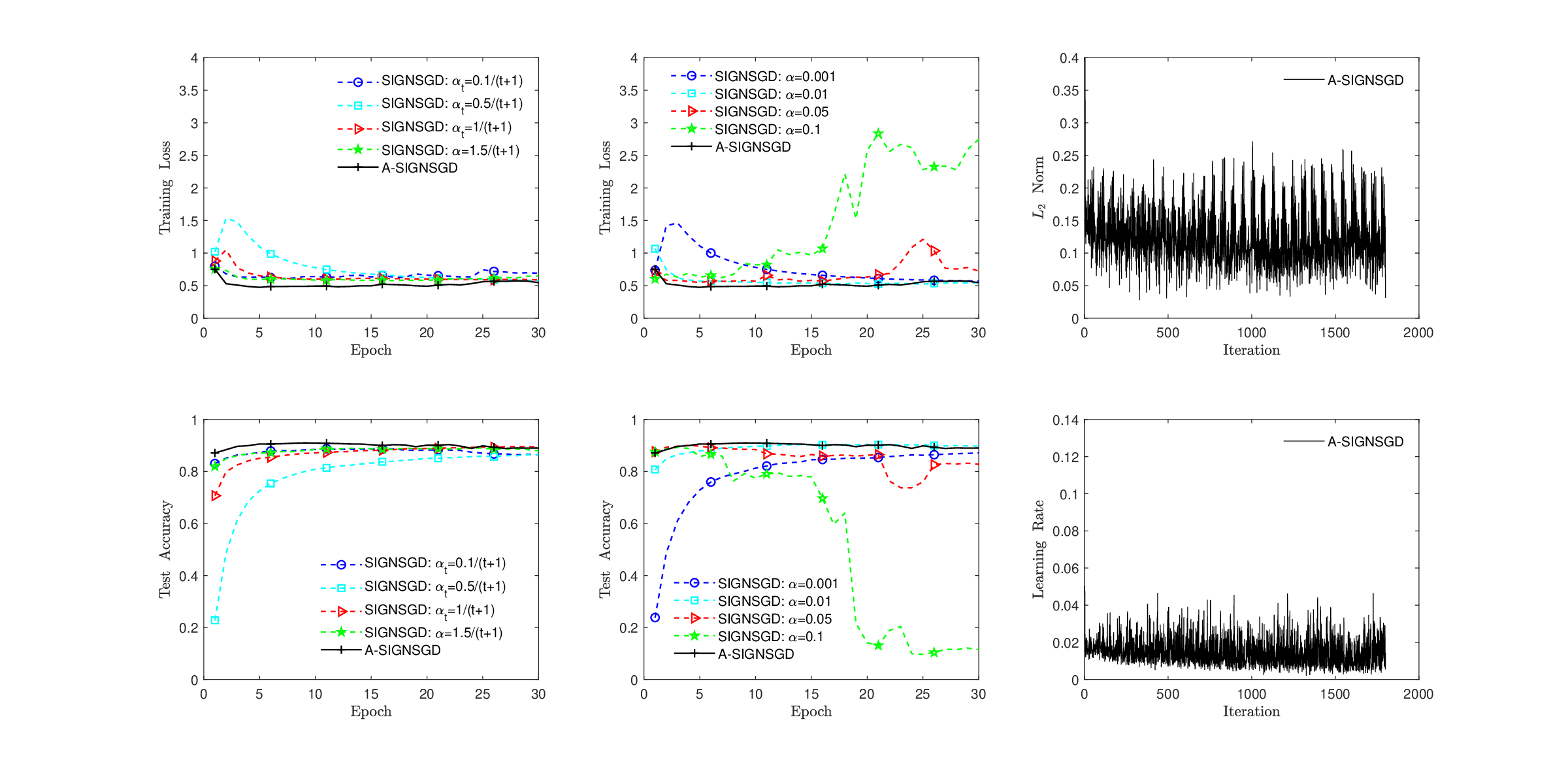}
\caption{MLP results on MNIST for A-SIGNSGD. The figure on the first and second columns plots the training loss comparison and classification accuracy of A-SIGNSGD and SIGNSGD with a diminishing LR sequence and different constant LRs in 30 epochs, respectively. The figure on the third column plots $\|\nabla F_{S_H}(x_t)\|$ and LRs of A-SIGNSGD every 10 iterations. The numbers in the legends are the LR, used in A-SIGNSGD. }\label{asignsgd-mlp}
\end{figure}

So as to better comprehend the numerical behavior of the algorithms with the adaptive LR scheme, Figs. \ref{asgd-mlp}, \ref{asgdm-mlp}, and \ref{asignsgd-mlp} display the performance of A-SGD, A-SGDM, and A-SIGNSGD and their original algorithms with a diminishing LR sequence and different constant LRs, respectively. In particular, we plot the training loss and the classification accuracy in the first and second columns of all figures, where in such the circumstance the $x$-axis denotes numbers of epochs. Each novel stochastic gradient evaluation $\nabla f_i(\cdot)$ counts as $1/m$ epoch and a complete evaluation $\nabla F(\cdot)$ counts as 1 epoch. In addition, we plot the $L_2$ norm,
 $\|\nabla F_{S_H}(x_t)\|$, and the LR in the third column of all figures, where the $x$-axis denotes the number of iterations under this case.

 The results in Figs. \ref{asgd-mlp}, \ref{asgdm-mlp}, and \ref{asignsgd-mlp} are discussed here. For SGD, SGDM, and SIGNSGD under mini-batching backgrounds, we first set a diminishing LR sequence as the rule $\alpha_t=c/(t+1)$, and then pick up a constant LR from the set \{0.001, 0.01, 0.05, 0.1\}, where the detailed information of $c$ is provided in the legend of figures. Note that it is difficult and impossible to show the performance of these algorithms with all cases of LRs. Therefore, as suggested by many studies, a better LR is selected by using the grid search technique from multiple LRs. For SGD, SGDM, and SIGNSGD, one epoch is denoted as $T=m/b$, while for A-SGD, A-SGDM, and A-SIGNSGD, one epoch is denoted as $T=m/(b+b_H)$ in order to compare fairly.

 We are now ready to describe the detailed information of Figs. \ref{asgd-mlp}, \ref{asgdm-mlp}, and \ref{asignsgd-mlp}.
\begin{enumerate}
\item[(i)] First of all, all figures demonstrate that the resulting algorithms converge faster than their original algorithms that use a diminishing LR regime. For the constant parameter $c$ in the original SGD algorithm, it is still required to be tuned manually for better performance. As a side note, the results presented in the first column of all figures are in accordance with the fact that a diminishing LR sequence leads to a slow convergence rate of stochastic algorithms.

 \item[(ii)] Further, all figures demonstrate that the resulting algorithms are comparable, and even perform better than their original algorithms with a well-calibrated LR. Additionally, Figs. \ref{asgdm-mlp} and \ref{asignsgd-mlp} indicate that a large LR makes SGDM and SIGNSGD attain worse performance. Even in practice, a large LR will lead to the divergence of stochastic optimization algorithms.

 \item[(iii)] Third, Figs.  \ref{asgd-mlp} and \ref{asgdm-mlp} illustrate that the LR sequence of A-SGD and A-SGDM is increasing as the iterative processes, which means that it is not necessary to make the stochastic algorithms work with a decreasing LR sequence to guarantee their convergence. This finding is consistent with the conclusion in \cite{umeda2025increasing,smith2018don}, but they ask for increasing batch sizes simultaneously to obtain a better performance of stochastic optimization algorithms.


 \item[(iv)] Finally, all figures indicate that the LR sequence increases when the norm of gradients decreases, or deceases when the norm of gradients increases. The numerical results in all figures show that the LR sequence dynamically varies along the variation of stochastic gradients. Regarding more specific reasons, it will be discussed below (see Section \ref{sec4}).
 \end{enumerate}

Besides, we have that the original SIGNSGD algorithm performs better under the case of $\alpha=0.001$.  Surprising, the LR sequence of A-SIGNSGD, shown in Fig. \ref{asignsgd-mlp}, oscillates around 0.001. The similar phenomenon also happens in A-SGD, where in A-SGD, the LR sequence is close to 0.05 at start time and over 0.05 later and gradually close to 0.1. Note that under the case of $\alpha=0.05$, or $\alpha=0.1$, the original SGD algorithm behaves better. Numerical results in Figs. \ref{asgd-mlp}, \ref{asgdm-mlp}, and \ref{asignsgd-mlp}, imply that the adaptive LR regime makes stochastic optimization algorithms attain an optimal LR automatically and implement it simply in practice without tuning any hyper-parameters (see later). Moreover, in contrast to the existing LR regimes, the resulting LR regime has a better interpretation, presenting how the LR sequence updates according to the variation of stochastic gradients.

\section{Discussion}\label{sec3}

Compared with the original SGD, SGDM, and SIGNSGD algorithms, A-SGD, A-SGDM, and A-SIGNSGD require additional computational expenditure to compute the LR per iteration step. It seems that the computational burden of the resulting algorithms is bigger than that of their original algorithms. We here theoretically analyze that the oracle complexity of A-SGD is comparable to that of SGD under mini-batching settings. For A-SGDM and A-SIGNSGD, we easily obtain a similar conclusion.

We are interested in a stochastic convex optimization problem, where we address the model $F(x)=\frac{1}{m}\sum_{i=1}^m f_i(x)$ over some convex domain $\mathcal{X}$. In order to better support our claim, we adopt similar assumptions to the work \cite{cotter2011better}. Throughout this work, we suppose that each component function $f_i(x)$ is convex, non-negative, and $H$-smooth. We also take $\mathcal{X}$ to be the set $\mathcal{X}=\{x: \|x\|\leq D\}$. Notice that SGD is an unbiased stochastic algorithm, i.e., $\E[\nabla f_i(x)]=\nabla F(x)$.

The following theorem establishes a convergence guarantee for A-SGD.

\begin{theorem}
\label{theo-1}
For A-SGD with $b_H=\max\{4H^2, \frac{64H^2}{b^2}\}$, supposing $F(x_0)\leq H D^2$, we have that
\begin{align}
\label{f-the-1}
\E[F(\bar{x})]-F(x_*)\leq \sqrt{\frac{128(H+1)D^2F(x_*)}{bn}}+\frac{2HD^2}{n}
+\frac{32(H+1)D^2}{bn},
\end{align}
where $\bar{x}=\frac{1}{n}\sum_{i=1}^n x_{t}$ and $F(x_*)$ is the expected loss of the optimal solution $x_*$.
\end{theorem}

For simplicity, the proof of Theorem \ref{theo-1} is given in Appendix. We are now discussing the theoretical results in Theorem \ref{theo-1}. Considering the iteration number $n=m/(b+b_H)$, we have the bound for A-SGD
\begin{align}
\label{f-the-2}
\E[F(\bar{x})]-F(x_*)\leq \tilde{O}\left(\sqrt{\frac{F(x_*)}{bn}}+\frac{1}{n}\right)
=\tilde{O}\left(\frac{(b+b_H)F(x_*)}{bm}+\frac{b+b_H}{m}\right).
\end{align}

The overall complexity of SGD with mini-batching, shown in \cite{cotter2011better}, is
\begin{align}
\label{f-the-3}
\E[F(\bar{x})]-F(x_*)\leq \tilde{O}\left(\sqrt{\frac{F(x_*)}{m}}+\frac{b}{m}\right).
\end{align}
The results in \eqref{f-the-2} and \eqref{f-the-3} show that the only difference between A-SGD and SGD in the oracle complexity is the introduction of the parameter $b_H$ in A-SGD, where we use $b_H$ samples to compute the LR. Actually, with small samples $S_H$, the complexity of A-SGD is comparable to the original SGD algorithm, which supports our claim that the oracle complexity of A-SGD is comparable to that of SGD. To show this, next, we will empirically confirm that the resulting algorithms are insensitive to $b_H$ .

In the following, we will see that the only factor that will influence our LR regime is the hyper-parameter $b_H$. Therefore, we here additionally discuss the effect of $b_H$ in the resulting algorithms. Concretely, we show the performance of A-SGD and A-SGDM with different $b_H$ on MLP, where $b_H$ is considered in the set \{10, 50, 100, 150\}.

\begin{figure}[!hptb]
\centering
\includegraphics{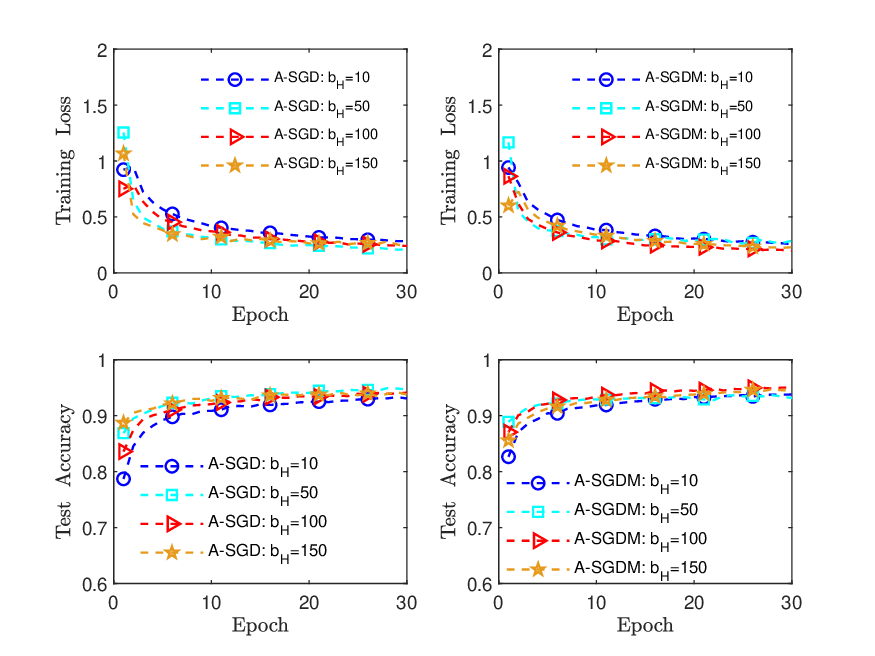}
\caption{Illustration of A-SGD (\textbf{left}) and A-SGDM (\textbf{right}) with different $b_H$ on MLP, where $b_H$ is a selection from the set \{10, 50, 100, 150\}. }\label{asgd-m-dbh}
\end{figure}

It is observed from Fig. \ref{asgd-m-dbh} that A-SGD and A-SGDM are almost insensitive to $b_H$, indicating the robustness of the resulting LR regime. The similar empirical results happen in A-SIGNSGD. As a result, we do not show the performance of A-SIGNSGD with different $b_H$ separately.

So far, we have theoretically and empirically discussed the positive role of the adaptive LR regime in stochastic optimization algorithms. The following will continually and clearly demonstrate how the variation of stochastic gradients impacts the LR sequence. Also, we will quantify the relationship between the LR and stochastic gradients soon.

\section{Methods}\label{sec4}

The idea of the LR regime, developed by this work, is motivated by the SQN method. The original SQN method \cite{byrd2016stochastic} for solving $\min_{x \in \R^d} F(x)=\frac{1}{m}\sum_{i=1}^m f_i(x)$ is formulated as $x_{t+1}=x_t-B_t^{-1}\nabla F_{S}(x_t)$, where $B_t=\nabla F_{S_H}(x_t)=\frac{1}{b_H}\sum_{i \in S_H} \nabla f_i(x_t)$ denotes an approximation of the Hessian matrix of the objective function $F(x)$ at point $x_t$ and $\nabla F_S(x_t)=\frac{1}{b}\sum_{i \in S} \nabla f_i(x)$ ($S\subset \{1, 2, \cdots, n\}$).

In the stochastic BB algorithm, one uses $B_t=\frac{1}{\alpha_t} I$ to approximate the real Hessian matrix. Further, by utilizing the secant equation $B_ts_t=y_t$, where $B_t=\eta I$, $s_t=x_t-x_{t-1}$ and $y_t=\nabla F_{S_H}(x_t)-\nabla F_{S_H}(x_{t-1})$, we obtain the BB-type learning rate. Nevertheless, unlike the BB-type algorithm, when dealing with the finite-sum problem $\min_{x \in \R^d} F(x)=\frac{1}{n}\sum_{i=1}^n f_i(x)$, we compute the LR by minimizing the residual of the secant equation, $\|B_ts_t-y_t\|^2$, where we set $\hat{s}_t=\nabla F_{S_H}(x_t)$, and $\hat{y}_t=\nabla F_{S_H}(x_t+\nabla F_{S_H}(x_t))-\nabla F_{S_H}(x_t)$. Therefore, we have a formula for computing the LR sequence of stochastic optimization algorithms, i.e.,
\begin{align}
\label{f-lr-our-1}
\alpha_t=\frac{\|\hat{s}_t\|^2}{\langle \hat{y}_t, \hat{s}_t \rangle}.
\end{align}
Notice that when adopting $\hat{s}_t=x_t-x_{t-1}$ and $\hat{y}_t=\nabla F_{S_H}(x_t)-\nabla F_{S_H}(x_{t-1})$ in \eqref{f-lr-our-1}, it becomes the stochastic BB-type learning rate. Additionally, when considering $\hat{s}_t=\nabla F(x_t)$, and $\hat{y}_t=\nabla F(x_t+\nabla F(x_t))-\nabla F(x_t)$ in \eqref{f-lr-our-1}, it is similar to the stochastic Steffensen learning rate, developed by \cite{zhao2024stochastic}.

Nevertheless, different from existing studies, this work considers the following LR regime
\begin{align}
\label{f-lr-our-2}
\alpha_t=\frac{1}{\sqrt{b_H}}\cdot\frac{\|\hat{s}_t\|^2}
{\langle \hat{y}_t, \hat{s}_t \rangle+\|\hat{s}_t\|^2}.
\end{align}
The iterative scheme of updating the LR \eqref{f-lr-our-2} inserts a term, $\|\hat{s}_t\|^2$, in the denominator of \eqref{f-lr-our-1}. Besides, we also multiply $\frac{1}{\sqrt{|S_H|}}$ in \eqref{f-lr-our-2} due to the use of mini-batching samples, $S_H \subset \{1, 2, \cdots, n\}$.


Here, we illustrate why the formula \eqref{f-lr-our-2} in acquiring the LR for stochastic algorithms is effective and interpretable. Intuitively, the LR sequence should increment when the update direction decreases, since at the optimal solution, $\lim_{t \to \infty}\|\nabla F(x_t)\|^2=0$; otherwise, it should be shrunk. The iterative regime \eqref{f-lr-our-2} is in accordance with this instinct spontaneously. Concretely, the iterative scheme \eqref{f-lr-our-2} can be further simplified to
\begin{align}
\label{f-lr-our-2-y1}
\alpha_t=\frac{1}{\sqrt{|S_H|}}\cdot\frac{\|\hat{s}_t\|^2}
{\langle \nabla F_{S_H}(x_t+\nabla F_{S_H}(x_t)), \nabla F_{S_H}(x_t) \rangle}.
\end{align}

The formulation \eqref{f-lr-our-2-y1} shows us that the LR sequence will increase when the model is towards the optimal solution. On the contrary, the LR sequence will decrease when the model is away from the optimal value. Considering the former case, from $\|\nabla F_{S_H}(x_t+\nabla F_{S_H}(x_t))\|\leq \|\nabla F_{S_H}(x_t)\|$, we ascertain that $\langle \nabla F_{S_H}(x_t+\nabla F_{S_H}(x_t)), \nabla F_{S_H}(x_t)\rangle\leq \|\nabla F_{S_H}(x_t)\|^2$. Further, we have $\frac{\|\hat{s}_t\|^2}{\langle \nabla F_{S_H}(x_t+\nabla F_{S_H}(x_t)), \nabla F_{S_H}(x_t) \rangle} >1$. More generally, the LR regime \eqref{f-lr-our-2} clarifies how LR can be updated automatically only according to the intrinsic variation of stochastic gradients.

To better comprehend the LR regime \eqref{f-lr-our-2}, the detailed comparison with the existing LR schemes is given here.
\begin{enumerate}[1.]
\item In contrast to the conventional LR rule for SGD, where the LR sequence should be decreased as $\alpha_t=\frac{c}{1+t}$, or $\alpha_t=\alpha_0(1+\gamma t)^{-1}$, the formula \eqref{f-lr-our-2} demonstrates that the LR sequence can increase, which makes it possible to accelerate stochastic optimization algorithms.

 \item Compared with several popular LR regimes for stochastic algorithms, such as the BB-type learning rate (see \eqref{f-5}), the Polyak learning rate (see \eqref{f-3}), Adam (see \eqref{f-1}), etc., the formula \eqref{f-lr-our-2} provides a better interpretation of how the LR automatically and intelligently varies only according to the gradient information along the iteration progress.

 \item Different from the existing heuristic algorithms and adaptive gradient methods, like a multiplicative LR modification rule, Adam, RMSProp, etc., where additional parameters are introduced, the LR regime \eqref{f-lr-our-2} is easily to be used in practice since it is not necessary to tune hyper-parameters.


\end{enumerate}

To confirm the positive role of the LR regime \eqref{f-lr-our-2} in stochastic optimization algorithms, we apply it to three popular stochastic optimization algorithms, SGD, SGDM, and SIGNSGD, leading to three novel stochastic optimization algorithms, coined A-SGD, A-SGDM, and A-SIGNSGD. For more details of these three algorithms, please see Algorithm \ref{alg-i}.

\begin{algorithm}[!hptb]
   \caption{\colorbox{green}{\textbf{A-SGD}} , \colorbox{lime}{\textbf{A-SGDM}} and \colorbox{pink}{\textbf{A-SIGNSGD}}}
   \label{alg-i}
\begin{algorithmic}[1]
 \State \textbf{Parameters:} mini-sample sizes $b$ and $b_H$
 \State \textbf{Input:} $x_0$ and $\beta$ (only for momentum-based algorithms)
   \For{$t=1$ {\bfseries to} $n=m/(b+b_H)$}
   \State Choose a sample $S_H \subset [n]$, with $|S_H|=b_H$ and compute the learning rate by the formula \eqref{f-lr-our-2}
   \begin{align*}
\alpha_t=\frac{1}{\sqrt{|S_H|}}\cdot\frac{\|\hat{s}_t\|^2}
{\langle \nabla F_{S_H}(x_t+\nabla F_{S_H}(x_t)), \nabla F_{S_H}(x_t) \rangle}.
\end{align*}
   \State Choose a sample $S \subset [n]$, with $|S|=b_H$
   and calculate stochastic gradient $\nabla F_S(x_t)$
\begin{align*}
 &\colorbox{green}{$x_{t}^{\mathrm{A-SGD}}=x_{t-1}-\alpha_t \nabla F_S(x_{t-1})$} \\
   &\colorbox{lime}{$x_{t}^{\mathrm{A-SGDM}}=x_{t-1}-\alpha_t \nabla F_S(x_{t-1})+\beta(x_t-x_{t-1})$} \\
 &\colorbox{pink}{$x_t^{\mathrm{A-SIGNSGD}}=x_{t-1}-\alpha_t \mathrm{ sign}(\nabla F_S(x_{t-1}))$}
 \end{align*}
   \EndFor
\State $\bar{x}=\frac{1}{n}\sum_{i=1}^n x_{t}$, or $\bar{x}=x_n$.
\end{algorithmic}
\end{algorithm}

\begin{remark}
Algorithm \ref{alg-i} varies among A-SGD, A-SGDM, and A-SIGNSGD by employing different update directions, such as in A-SGD, $d_k=\nabla F_S(x_k)$; in A-SIGNSGD, $d_k=\mathrm{sign} (\nabla F_S(x_k))$. More generally, the LR regime \eqref{f-lr-our-2} can be readily incorporated into other stochastic optimization algorithms, such as stochastic conjugate gradient (SCG) \cite{yang2023adaptiveieee}, stochastic second-order algorithms \cite{agarwal2017second}, stochastic mirror descent \cite{paul2025convergence}, etc. This work does not intend to show the case of these stochastic optimization algorithms with adaptive LR regime \eqref{f-lr-our-2}.
\end{remark}

In the above experiments on MLP, we have the idea about how the LR sequence and the stochastic gradient vary for different algorithms and roughly establish the relationship between LR and stochastic gradients. It seems that the LR sequence is automatically increasing or decreasing with respect to the change of stochastic gradients. Concretely, the LR sequence increases when the norm of stochastic gradients decreases and the LR sequence decreases when the norm of stochastic gradients increases. To further confirm this claim, we perform the resulting algorithms on both the logistic regression model and the support vector machine (SVM) model.

\begin{figure}[!hptb]
\centering
\subfigure[\label{fis-a}][A-SGD on $a8a$ for LR]{\includegraphics[width=6.4cm]{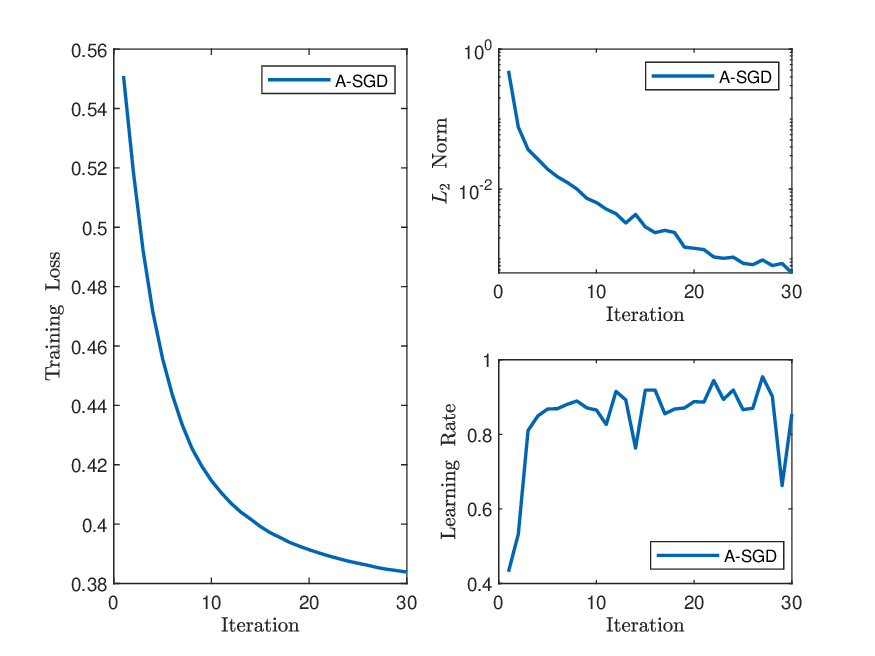}}
\hspace{-4mm}
\subfigure[\label{fis-a}][A-SGD on $covtype$ for LR]{\includegraphics[width=6.4cm]{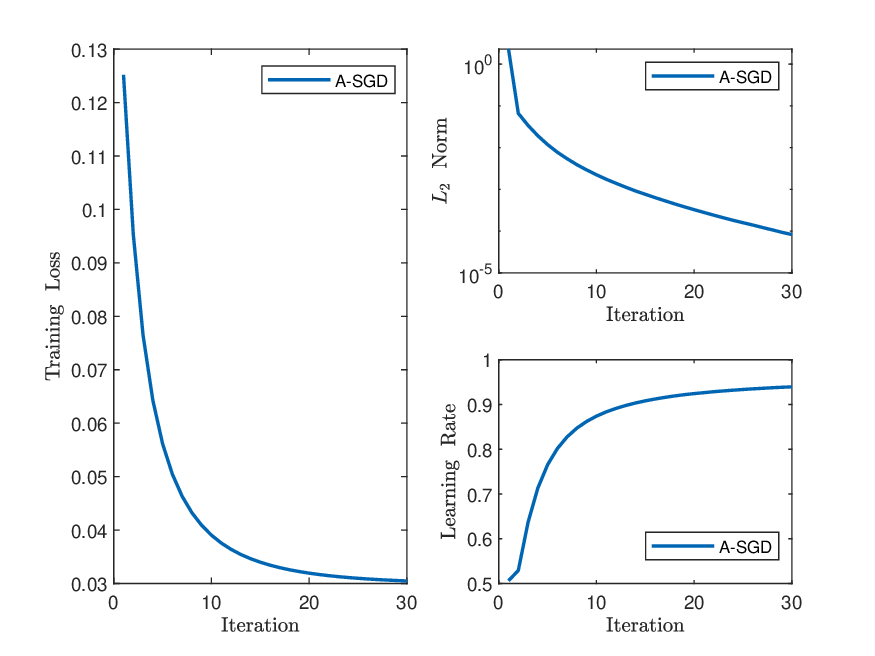}}\\
\subfigure[\label{fis-a}][A-SGD on $a8a$ for SVM]{\includegraphics[width=6.4cm]{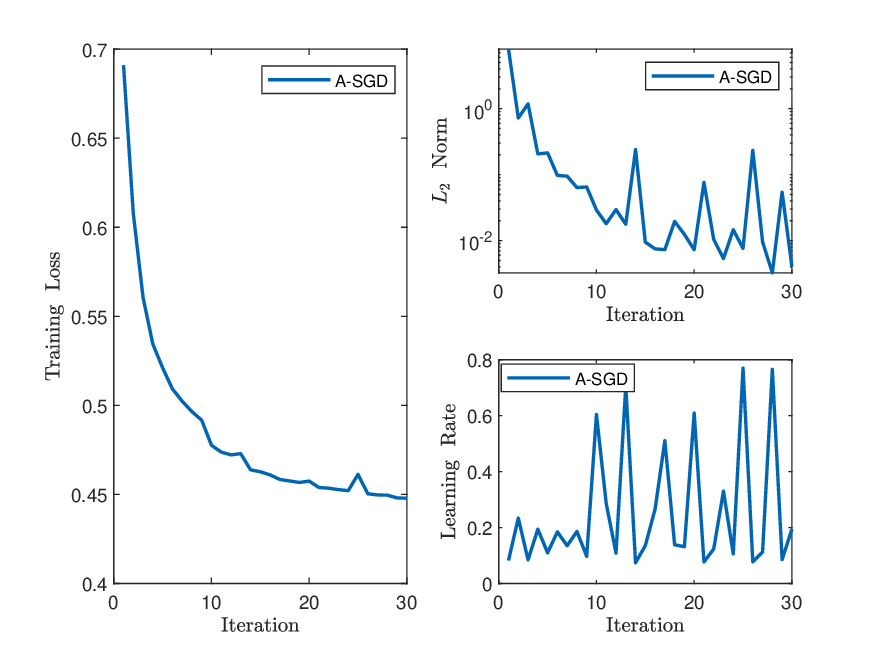}}
\hspace{-4mm}
\subfigure[\label{fis-a}][A-SGD on $covtype$ for SVM]{\includegraphics[width=6.4cm]{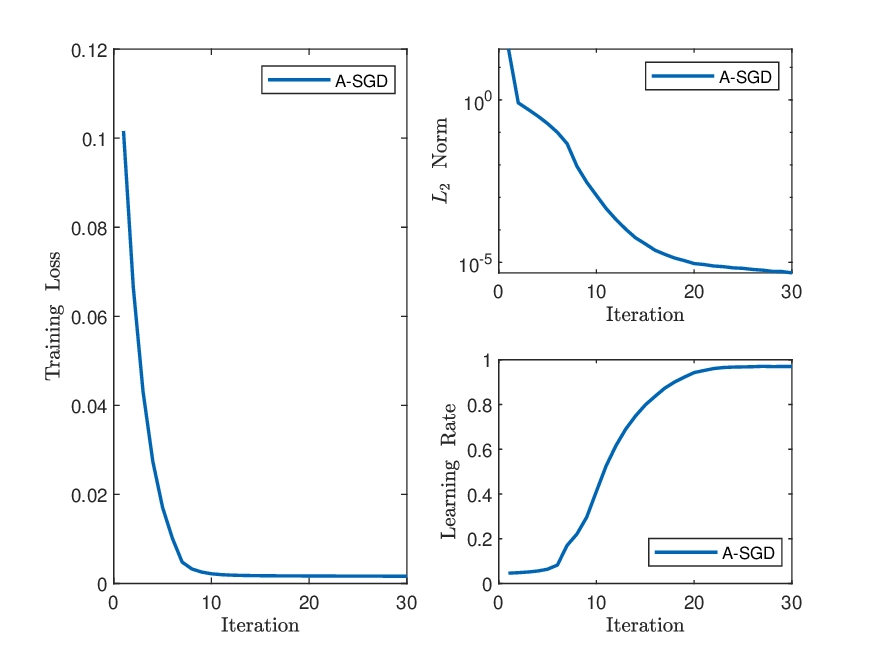}}\\
\caption{Logistic regression (\textbf{top}) and support vector machine (\textbf{bottom}) with A-SGD on $a8a$ (\textbf{left}) and $covtype$ (\textbf{right}). In the left figures of Figs. (a) and (b), we plot the training loss. In the right figures of Figs. (a) and (b), we plot the $\|\nabla F(x)\|$ (top) and the learning rate (bottom). }
\label{fig:pro-a1-lr-test}
\end{figure}

First, Fig. \ref{fig:pro-a1-lr-test} plots the training loss, along with the norm of gradients and the LR sequence for A-SGD on logistic regression (the top line) and SVM (the bottom line) with two publicly available datasets from LIBSVM \cite{chang2011libsvm} (a.k.a. $a8a$ and $covtype$), respectively. To better comprehend the connection between the gradient and the learning rate, we will take $b=b_H=n$ for different algorithms. Observed from Fig. \ref{fig:pro-a1-lr-test}, the decrease of the norm of the gradient leads to the increase of the LR. More specifically, the change of the LR sequence accurately follows from the variation of the gradient. In other words, the LR regime \eqref{f-lr-our-2} makes A-SGD automatically obtain the LR according to the gradient of the objective function intrinsically without requiring other information. Notice that when plotting the norm of gradients, we take $\log \|\nabla F(x_k)\|$ as a sign, which can more clearly exhibit the minor change of gradient information. Actually, if we directly take $\|\nabla F(x_k)\|$ as an indication, we will see that $\|\nabla F(x_k)\|$ is decreasing fluently, where it is not easy to show the relationship between the variation of stochastic gradients and the LR.

\begin{figure}[!hptb]
\centering
\subfigure[\label{fis-a}][A-SGDM on $a8a$ for LR]{\includegraphics[width=6.4cm]{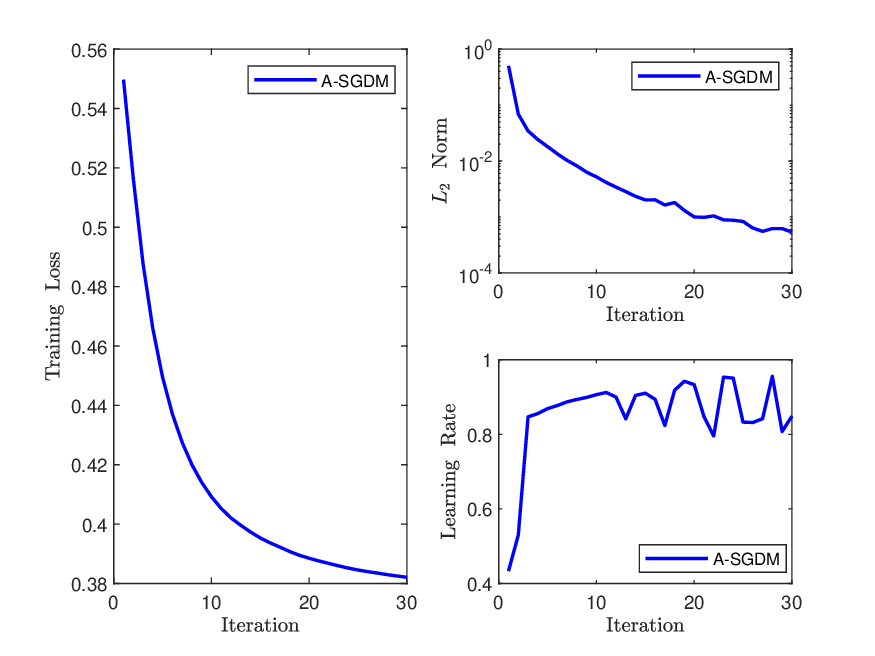}}
\hspace{-4mm}
\subfigure[\label{fis-a}][A-SGDM on $covtype$ for LR]{\includegraphics[width=6.4cm]{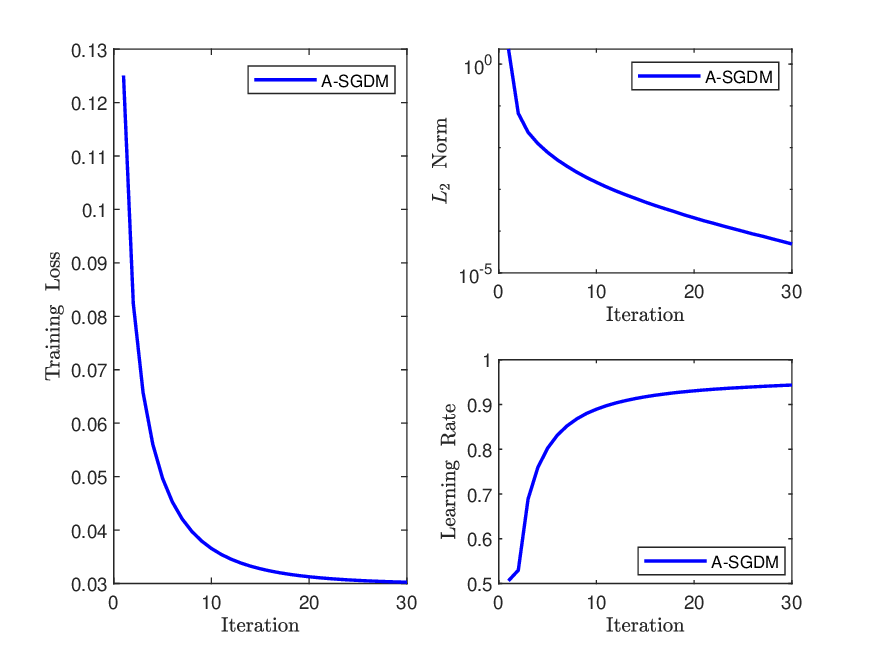}}\\
\subfigure[\label{fis-a}][A-SGDM on $a8a$ for SVM]{\includegraphics[width=6.4cm]{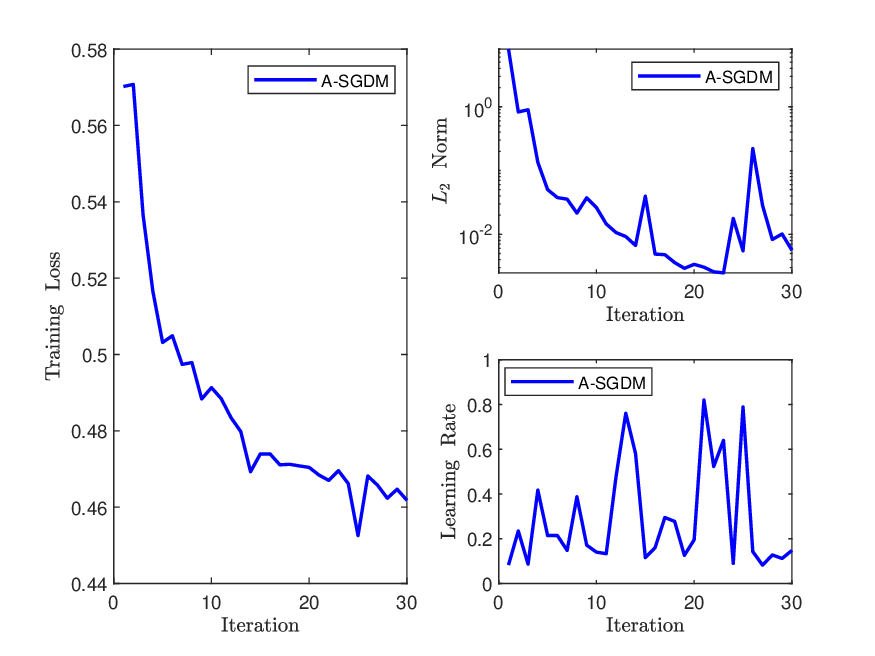}}
\hspace{-4mm}
\subfigure[\label{fis-a}][A-SGDM on $covtype$ for SVM]{\includegraphics[width=6.4cm]{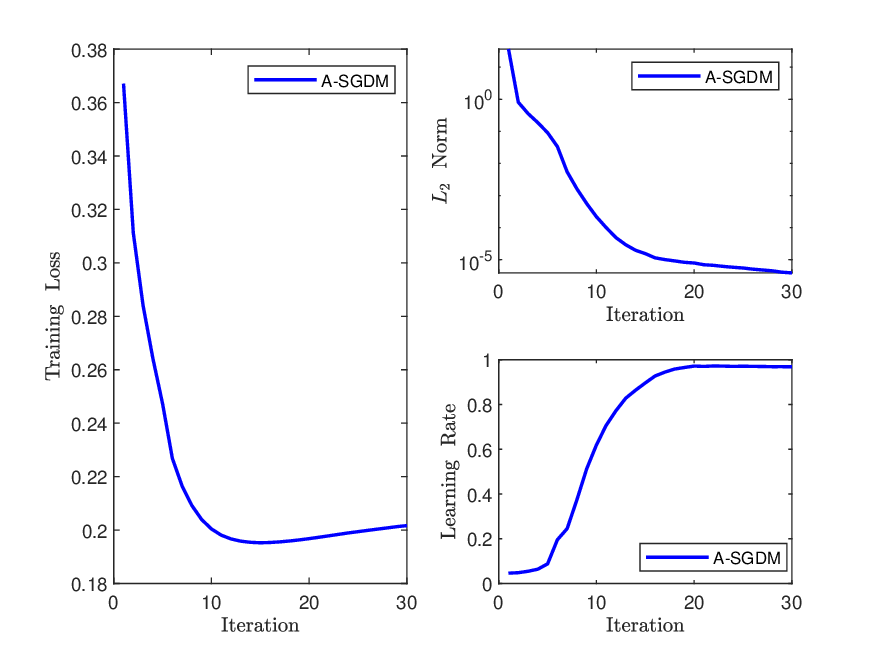}}\\
\caption{Logistic regression (\textbf{top}) and support vector machine (\textbf{bottom}) with A-SGDM on $a8a$ (\textbf{left}) and $covtype$ (\textbf{right}). In the left figures of Figs. (a) and (b), we plot the training loss. In the right figures of Figs. (a) and (b), we plot the $\|\nabla F(x)\|$ (top) and the learning rate (bottom). }
\label{fig:pro-a2-lr-test}
\end{figure}

Further, Fig. \ref{fig:pro-a2-lr-test} plots the training loss, along with the norm of gradients and the LR for A-SGDM on both logistic regression (the top line) and SVM (the bottom line) with $a8a$ and $covtype$, respectively. The setting for A-SGDM is similar to A-SGD, but the momentum coefficient is $\beta=0.9$, specifically for momentum-based algorithms. As seen from Fig. \ref{fig:pro-a2-lr-test}, A-SGDM performs similarly to A-SGD on different datasets.

All results in Figs. \ref{asgd-mlp}-\ref{asignsgd-mlp} and Figs. \ref{fig:pro-a1-lr-test}-\ref{fig:pro-a2-lr-test} demonstrate the LR regime \eqref{f-lr-our-2} makes the algorithms automatically pick up appropriate, or even optimal LRs based on the change of the gradient information. More specifically, the update rule of the LR in \eqref{f-lr-our-2} does not ask the LR sequence of stochastic optimization algorithms to decrease as the algorithms progress. More importantly, compared with the existing LR regimes, the LR regime \eqref{f-lr-our-2} is more interpretable for stochastic optimization algorithms, and specifically requires much less hyper-parameter tuning, which will be more suitable for different practical applications.

\backmatter

%
%
%

\bmhead{Acknowledgements}

This work was supported by grants from the National Natural Science Foundation of China under Grant 62302325. This work was also supported by the Natural Science Foundation of Jiangsu Province in China under Grant BK20230485 and by Project Funded by the Priority Academic Program Development of Jiangsu Higher Education Institutions.

\section*{Statements and Declarations}
\subsection*{\bf{Conflict of interest}}
Author Zhuang Yang declares that he has no conflict of interest.

\subsection*{\bf{Ethical approval}}
This article does not contain any studies with human participants or animals performed by any of the authors.

\subsection*{\bf{Informed consent}}
Additional informed consent is obtained from all individual participants for whom identifying information is included in this article.

\subsection*{\bf{Authorship contribution statement}}
Zhuang Yang: Data curation, Formal analysis, Investigation, Methodology, Validation, Visualization, Writing - original draft, Funding acquisition.

%
%


\bibliography{sn-bibliography}




\end{document}